\documentclass[pre,twocolumn,twoside,superscriptaddress,floatfix]{revtex4-1}

\usepackage{natbib}

\usepackage{listings, color}

\usepackage{graphicx,epsfig,verbatim,enumerate}
\usepackage{amssymb,amsmath,amsthm,amsbsy}
\usepackage{ifthen}
\usepackage{morefloats}

\usepackage{algorithm}
\usepackage{algpseudocode} 
\usepackage{url}
 
\usepackage{longtable}
\usepackage{booktabs}
\usepackage{tabularx}
\usepackage{longtable}
\usepackage{hyperref} 
\hypersetup{colorlinks=true, linkcolor=blue, citecolor=blue}

\usepackage{textcomp}

\usepackage{mathtools}
\usepackage{bm}

\newboolean{twocolswitch}

\newcommand{\sindex}[1]{}
\newcommand{\nindex}[1]{}

\newcommand{\www}[1]{\url{#1}}

\usepackage{lettrine}

\newcommand{\pbid}[1]{p_{\text{bid},#1}}

\newcommand{\dvb}{\Delta V^{(b)}}
\newcommand{\dva}{\Delta V^{(a)}}
\newcommand{\dvbhat}{\Delta\hat{V}^{(b)}}
\newcommand{\vahat}{\hat{V}^{(a)}}
\newcommand{\vbhat}{\hat{V}^{(b)}}
\newcommand{\dvahat}{\Delta\hat{V}^{(a)}}
\newcommand{\dx}{\Delta X}
\newcommand{\phat}{\hat{p}}
\newcommand{\dcash}{\Delta c}

\newcommand{\shares}{N}
\newcommand{\dshares}{\Delta N}
\newcommand{\dprofit}{\Delta \pi}
\newcommand{\side}{s}
\newcommand{\ordershares}{N^{(o)}}
\newcommand{\orderprice}{X^{(o)}}
\newcommand{\dprice}{\Delta X^{(o)}}
\newcommand{\fmarg}{f_{\text{marg}}}
\newcommand{\generative}{\mathcal{G}}
\newcommand{\zi}{\text{ZI}}
\newcommand{\zip}{\text{ZIP}}
\newcommand{\mo}{\text{MO}}
\newcommand{\mr}{\text{MR}}
\newcommand{\mm}{\text{MM}}
\newcommand{\fv}{\text{FV}}
\newcommand{\nn}{\text{NN}}
\newcommand{\mechanism}{\mathcal{M}}
\newcommand{\agent}{\mathcal{A}}

\definecolor{mygreen}{rgb}{0,0.6,0}
\definecolor{mygray}{rgb}{0.5,0.5,0.5}
\definecolor{mymauve}{rgb}{0.58,0,0.82}

\begin{document}
\lstset{language=Python} 
\title{Evolving \textit{ab initio} trading strategies in heterogeneous environments}

\author{\firstname{David Rushing }\surname{Dewhurst}}
\thanks{david.dewhurst@uvm.edu}
\affiliation{
	MassMutual Center of Excellence in Complex Systems and Data Science,
	The University of Vermont,
	Burlington, VT 05405
  }
\affiliation{
  Vermont Complex Systems Center,
  Computational Story Lab,
  and Department of Mathematics and Statistics,
  The University of Vermont,
  Burlington, VT 05405
  }

\author{\firstname{Yi} \surname{Li}}
\affiliation{
	MassMutual Data Science,
	Boston, MA 02210
}
\affiliation{
	MassMutual Center of Excellence in Complex Systems and Data Science,
	The University of Vermont,
	Burlington, VT 05405
  }

\author{\firstname{Alexander} \surname{Bogdan}}
\affiliation{
	MassMutual Data Science,
	Boston, MA 02210
}
\affiliation{
	MassMutual Center of Excellence in Complex Systems and Data Science,
	The University of Vermont,
	Burlington, VT 05405
  }

\author{\firstname{Jasmine} \surname{Geng}}
\affiliation{
	MassMutual Data Science,
	Boston, MA 02210
}
\affiliation{
	MassMutual Center of Excellence in Complex Systems and Data Science,
	The University of Vermont,
	Burlington, VT 05405
  }

\begin{abstract}
Securities markets are quintessential complex adaptive systems in which heterogeneous agents compete in an attempt to maximize returns.
Species of trading agents are also subject to evolutionary pressure as entire classes of strategies become obsolete and new classes emerge.
Using an agent-based model of interacting heterogeneous agents as a flexible environment that can endogenously model many diverse market conditions,
we subject deep neural networks to evolutionary pressure to create dominant trading agents.
After analyzing the performance of these agents and noting the emergence of anomalous superdiffusion through the evolutionary process,
we construct a method to turn high-fitness agents into trading algorithms. We backtest these trading algorithms on real high-frequency foreign exchange data, demonstrating that elite trading algorithms are consistently profitable in a variety of market conditions---even though these algorithms had never before been exposed to real financial data.
These results provide evidence to suggest that developing \textit{ab initio} trading strategies by repeated simulation and evolution in a mechanistic market model may be a practical alternative to explicitly training models with past observed market data.
\end{abstract}

\maketitle

\section{Introduction}
\textit{Ab initio} artificial intelligence---algorithms that are capable of learning or evolving master-level performance from a zero-knowledge baseline on a task normally performed by humans---is a long-held goal of the field in general \cite{mnih2015human}.
Recent years have seen substantial progress toward this goal
\cite{mnih2015human,silver2017mastering,silver2018general}.
One particular area of interest is the development of algorithms that are able to trade financial assets without human supervision. 
The difficulty of this problem is enhanced by its fundamentally stochastic nature unlike the deterministic non-cooperative games of Go, chess, shogi, and Atari.
of obvious economic incentives for intense competition amongst candidate solution algorithms: if an algorithm has a non-transient ability to make a statistically significant positive profit, the owner of that algorithm stands to reap large financial gains.

Though there has been prior work on \textit{ab initio} trading strategies, such work has focused on small,
homogeneous collections of agents that interact over shorter timescales than those considered in this study
\cite{chen2009co}.
Trading strategies that use statistical and algorithmic methods more broadly are exceptionally common in the quantitative finance literature \cite{garcia2015social,golub2018alpha,greenwald20032002},
and are used in practice with mixed results
\cite{davidson2012dark,kirilenko2013moore,kissell2005understanding}.
\begin{figure*}
    \centering
    \includegraphics[width=0.9\textwidth]{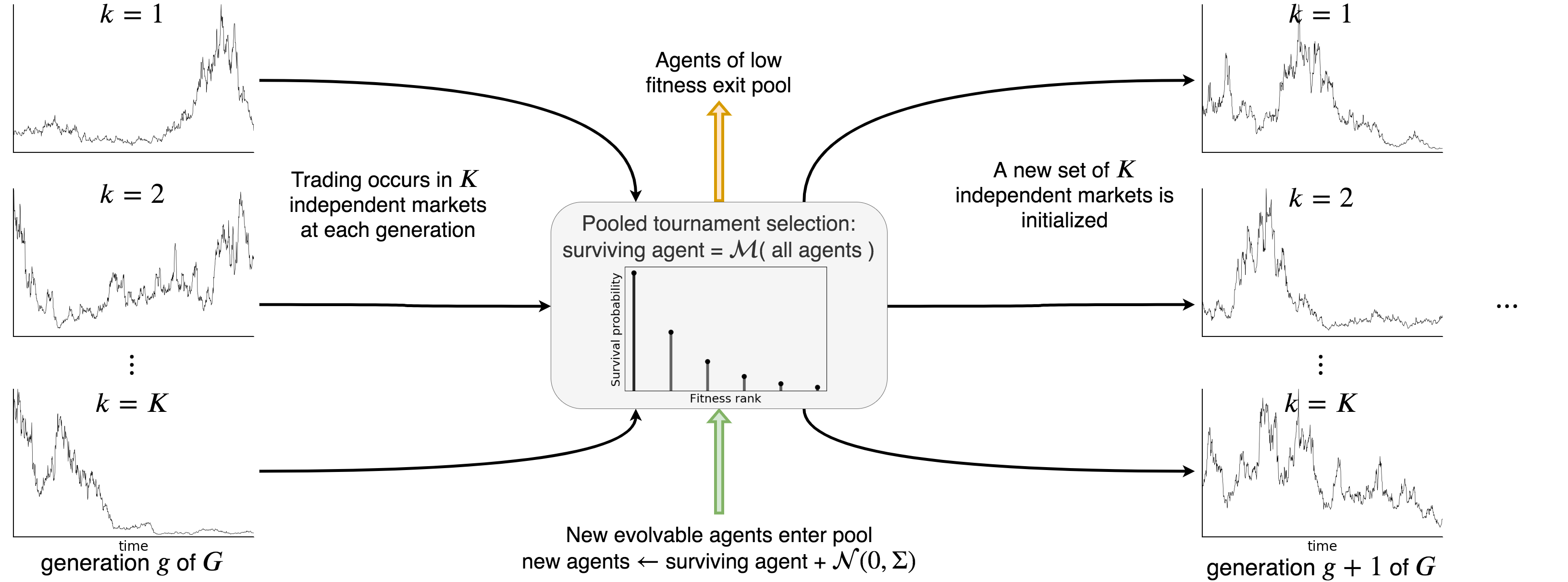}
    \caption{
    At each generation $g$ of the evolutionary process, we initialize $K$ independent markets in which agents (Sec.\ \ref{sec:agent-description}) interact via the matching engine described in Sec.\ \ref{sec:auction}. 
    At the end of $T$ timesteps, agents subject to evolution are 
    pooled, selection is applied, and then new agents are introduced using the mechanism described in Sec. \ref{sec:evo-dyn}. The process then begins again in generation $g + 1$ and continues for a total of $G$ generations.
    }
    \label{fig:mech-graphic}
\end{figure*}
Evolutionary approaches to the development of trading strategies have focused on derivation
of technical trading rules using observed market data as the training dataset and then backtesting the evolved rules on out-of-sample test data.
\cite{dempster2001real,austin2004adaptive}.
Likewise, evolutionary computation and agent-based models have been used extensively to study the macro properties of artificial asset markets rather than specifically studying the micro properties of individual trading strategies
\cite{lebaron1999time,palmer1999artificial,lebaron2002building,chen2001evolving,cincotti2003wins,cont2007volatility,martinez2009heterogeneous}. 
However, to our knowledge there has been no academic study of the possibility of developing \textit{in vivo} trading strategies using purely \textit{ab initio} methods---trading strategies that train or evolve using artificial data only, and then, at test time, trade using real asset prices---which is the approach that we take here. 

We pursue this objective for two reasons: first, achieving this goal would be a useful step in the development of evolutionary ``self-play'' techniques in the context of stochastic games with many players
\cite{runarsson2005coevolution,akchurina2009multiagent,heinrich2015fictitious}; 
and second, this would demonstrate that the development of profitable trading strategies could be realized by attempting to simulate with increasing accuracy the underlying mechanisms of financial markets instead of by predicting future real market prices.

The paper proceeds as follows: in Sec.\ \ref{sec:theory}, we 
describe the theory and details behind our agent-based financial market model, including the design of the price-discovery (auction) mechanism, heterogeneous agent population, evolutionary algorithm (summarized graphically in Fig.\ \ref{fig:mech-graphic}), and method to convert evolved individuals into trading strategies;
in Sec.\ \ref{sec:results} we summarize descriptive and quantitative results of the evolutionary dynamics and the performance of evolved trading algorithms backtested on real data; and in Sec.\ \ref{sec:discussion} we discuss these results and provide suggestions for future work. 

\section{Theory and simulation}\label{sec:theory}
Our simulation methodology is based on an agent-based market model (ABM) composed of a matching engine and heterogeneous agents, which we describe presently
\footnote{All source used in this project is available from the authors upon reasonable request.}. We then outline the evolutionary mechanism, how it interfaces with the ABM, and the methodology by which we generate functional trading strategies from evolved agents. 

\begin{figure*}[!htp]
    \centering
    \includegraphics[width=\textwidth]{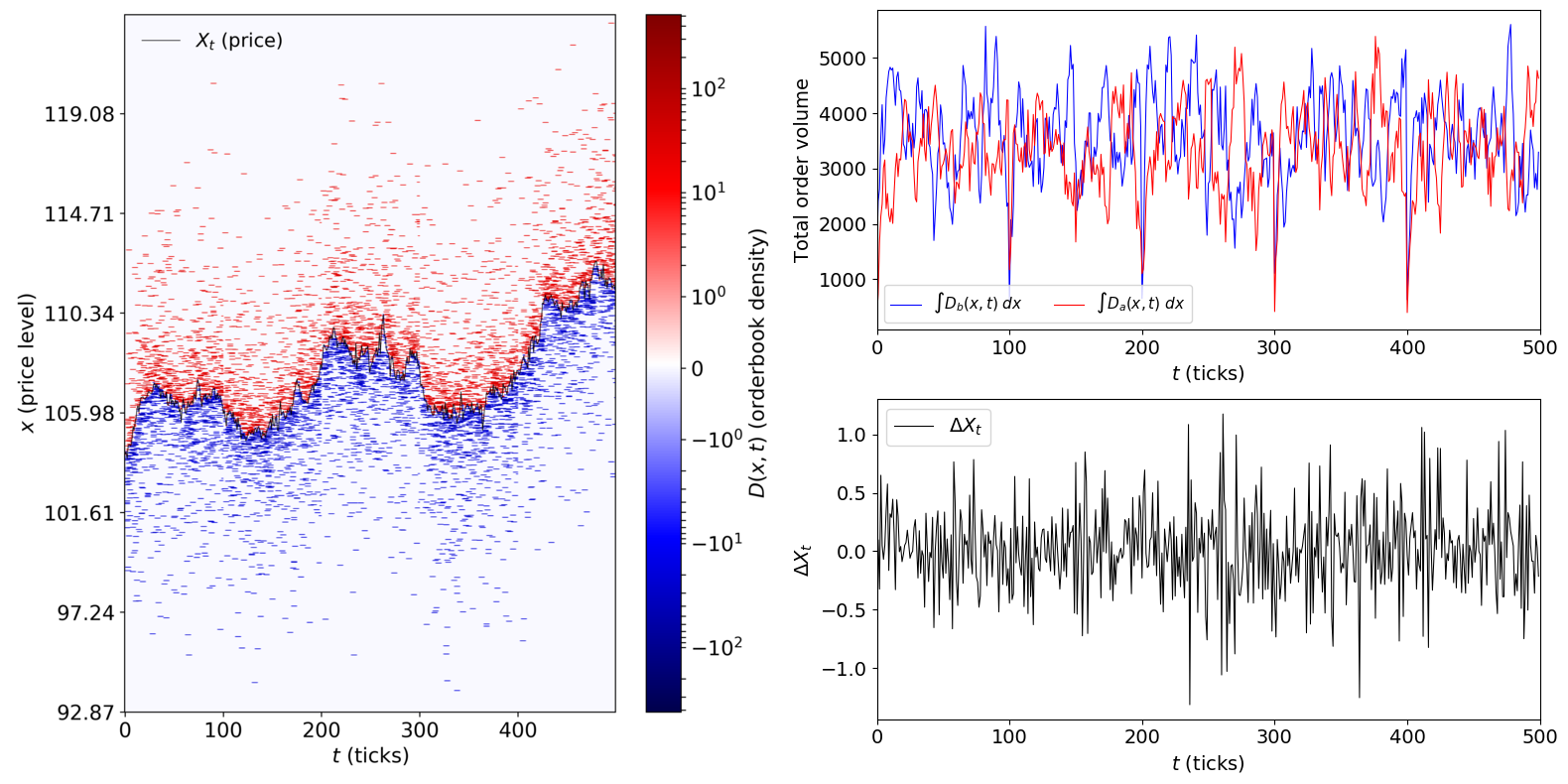}
    \caption{
    In panel A, we display an example orderbook corresponding with a very simple market simulation along with the resulting asset price time series. 
    We denote bid interest (plotted in blue) by negative numbers (corresponding with positive $D_b(x,t)$ later) and ask interest (plotted in red) by positive numbers 
    (corresponding with positive $D_a(x,t)$ later).
    We display the time series of total bid and ask interest in panel B and the time series of $\dx$ in panel C. Differences in total bid and ask interest are strongly associated with changes in level of $\dx$.
    The periodicity of large drops in $\int D_b(x, t)\ dx$ and 
    $\int D_a(x, t)\ dx$ in panel B is due to end-of-day orderbook clearing by the matching engine.
    }
    \label{fig:orderbook}
\end{figure*}
\subsection{Details of price discovery mechanism}\label{sec:auction}

At each timestep $t$, agents can submit orders to the matching engine, which attempts both to find an equilibrium price for that timestep and to match orders with one another so that exchange of shares for cash can occur. 
Orders are described by a three-tuple, $o = (\side, \ordershares, \orderprice)$, consisting of 
the desired side $s \in \{ \text{buy},\ \text{sell}\}$,
the number of shares that the agent would like to purchase or sell $\ordershares$,
and the requested price at which the agent would like to transact $\orderprice$.
The matching engine collects all orders submitted to it and matches bid orders with ask orders using a 
frequent batch auction (FBA) mechanism
\cite{budish2014implementation,budish2015high}.
This mechanism is an alternative to the continuous double auction (CDA) mechanism that is in use in
most securities markets. 
Both CDAs and FBAs are double-sided auctions, meaning that they match multiple buyers with multiple sellers 
of an asset at the same time, unlike auctions that match a single seller with multiple buyers 
(e.g., English, Japanese, or Dutch auctions) \cite{milgrom1982theory,mcafee1987auctions,ausubel2004efficient}.
However, CDAs and FBAs differ fundamentally as CDAs operate in continuous time and FBAs operate in discrete time. 
CDAs match orders as they are received; if the order cannot be immediately executed, it is placed into an order
 book where it waits to be matched with a future incoming order. 
In contrast, an FBA collects orders in discrete time trading intervals. At the end of each trading interval, 
orders are sorted according to price preference (bids are sorted from highest to lowest price, while asks are sorted from lowest to highest).
Matching then occurs in price-time priority order, meaning that bids with a higher price and asks with a lower
price are matched first. 
Ties in price are broken by age, where older orders---orders that are already resting in the orderbook from 
previous batches---are matched first 
\footnote{
The price discovery algorithm used in our matching engine implementation is a modified version of the logic 
used in the Australian Securities Exchange's matching engine \cite{comerton2006influence}. 
An open-source implementation of the matching engine is at 
\href{https://gitlab.com/daviddewhurst/verdantcurve}{https://gitlab.com/daviddewhurst/verdantcurve}.
}.
Orders submitted at timestep $t$ that do not execute at timestep $t$ remain in the orderbook for
consideration in future time periods.
Orders that remain in the orderbook---and hence are not matched with new, arriving orders---past 
a certain amount of time are considered ``stale'' by the matching engine and are removed. 
In our implementation, we set this period of time to be one day (or 100 time increments).
Agents are able to observe the price $X_t$ at time $t$ and the volume of resting bid ($b$) and ask ($a$) orders in the orderbook at price level $x$ at time $t$, written $D_b(x, t)$ and $D_a(x, t)$. 
All statistics used by agents in calculating side of book, price level and number of shares to submit are functions of these observable random variables and of the agents' own internal state, which we describe in the next section.
We display an example orderbook, along with the corresponding price trajectory, in panel A of Fig.\ \ref{fig:orderbook}.

\subsection{Heterogeneous agents}\label{sec:agent-description}

Our agent population is heterogeneous, comprised of seven qualitatively distinct varieties (species) of agents that analyze statistical behavior of asset prices differently and concomitantly exhibit divergent trading behavior.
We do not apply evolutionary pressure to six out of seven of these species; we thus separate the descriptions of the agents below into those not subject to evolutionary pressure (static agents) and those that do evolve.

\subsubsection{Static agents}
There are six types of agents in our model to which we do not apply evolutionary pressure. We give a brief overview of them below (the curious reader is encouraged to consult the supplementary information for more detail).
\begin{itemize}
    \item \textit{Zero-intelligence} ($\zi$): Inspired by work on statistical aspects of asset markets \cite{gode1993allocative,farmer2005predictive}, these agents submit a random bid or ask order with order price uniformly distributed around the last market price and number of shares Poisson-distributed around a fixed value (here set to be 100 shares).
    \item \textit{Zero-intelligence priceless} ($\zip$): identical to zero-intelligence agents, except these agents submit ``market'' orders --- orders that do not have a price but rather have first-priority execution and execute against the highest bid (for a market ask) or lowest ask (for a market bid).
    \item \textit{Momentum} ($\mo$): Momentum trading agents postulate that prices that have recently risen will continue to rise (and that prices that have recently fallen will continue to fall) \cite{chan1996momentum,badrinath2002momentum}. Our agents submit bid orders if the change in price is above some static positive threshold and ask orders if the change in price is below a symmetric negative threshold.
    \item \textit{Mean-reverting} ($\mr$): these agents postulate a return to some mean value of the asset price
    \cite{de1989anomalies,zhang2008trading}. When the asset prices move above a rolling mean value, these agents submit ask orders; when the price moves below the rolling mean, they submit bid orders.
    \item \textit{Market-making} ($\mm$): market-making strategies attempt to profit off of small price imbalances on either side of the orderbook; in doing so, they provide liquidity to the asset market
    \cite{o1986microeconomics,das2008effects}.
    \item \textit{Fundamental-value} ($\fv$): these agents have a certain fixed price set at which they ``believe'' the asset is fairly valued; if the asset price rises above that fundamental value, they submit ask orders, while if the asset price falls below it, they submit bid orders. 
    Note that this approach differs from that of $\mr$ traders since $\fv$ traders' valuations of the asset do not change over time.
\end{itemize}
We give more details on the implementation of these strategies in the supplementary information.
Though this typology of strategies has significant overlap with that introduced in the context of modern-day futures markets \cite{kirilenko2017flash}, it also exhibits some differences --- in particular, we do not implement a pure ``high-frequency trader'' agent since this does not make sense in the context of an FBA
\cite{budish2015high,aldrich2018experiments}.
We implement such a wide variety of strategies so that, in order for evolving agents to achieve high fitness, they must perform well in many different environments. We believe this will increase the likelihood of high-fitness agents performing well when confronted with real price data on which to trade, since real asset markets are also composed of heterogeneous agents \cite{kirilenko2017flash}.

\subsubsection{Evolvable agents}
We model technologically-advanced agents that are subject to evolutionary pressure as deep neural networks, denoted by 
$f_\theta$ (we will omit the vector of parameters $\theta$
when it is contextually unnecessary).
These neural networks take as inputs changes in price, changes in total orderbook volume, and changes in internal state (cash, shares held, and profit) and output a three-tuple of side (bid or ask), number of shares in the order, and price level of the order:
\begin{equation}\label{eq:nn-definition}
    (\side_t, \ordershares_t, \dprice_t) = 
    f(\dx, \dvbhat, \dvahat, \dcash, \dshares, \dprofit)
\end{equation}
We briefly describe the inputs of this function and their derivation from observable market statistics.
The change in asset price from $t-1$ to $t$ is given by 
$\dx_t = X_t - X_{t-1}$, while the change in cash ($\dcash$), shares ($\dshares$), and profit ($\dprofit$) are defined analogously.
The total bid and ask interest in the orderbook at time $t$ are given by $
\vbhat_t = \int_0^{X_t}D_b(x, t)\ dx $
and $\vahat_t = \int_{X_t}^{\infty} D_a(x, t)\ dx$
respectively; we then have 
$\dvbhat_t = \vbhat_t - \vbhat_{t-1}$ 
and 
$\dvahat_t = \vahat_t - \vahat_{t-1}$.
The neural network is a four-layer feed-forward model.
The two hidden layers have 20 and 10 neurons respectively, for a total of 383 evolvable parameters in each neural network
\footnote{
Number of parameters is equal to number of parameters in the weight matrices plus the number of parameters in the bias vectors 
$= 350 + 33$.
}.

It is an analytical convenience to consider a single generation of the evolutionary process described in 
Sec.\ \ref{sec:evo-dyn} as a draw from a stochastic function 
$\generative(\alpha, \mechanism)$,
where $\alpha$ describes the specification of the agents in the model and $\mechanism$ is the evolutionary mechanism (selection and mutation); we will describe these parameters in some detail in Sec.\ \ref{sec:evo-dyn}.
This function yields orderbooks and price time series; each call to the function yields a tuple of bid and ask order density and a price time series, 
\begin{equation}\label{eq:gen-mech}
(D_b(x, t), D_a(x, t), X_t) = \generative(\alpha, \mechanism).
\end{equation}
This way of looking at the process makes it easy to express Monte Carlo estimates of theoretical quantities and provides the theoretical basis for conversion of evolved agents into trading algorithms as we outline in Sec. \ref{sec:trading-algo-description}.
We can re-express the above integrals as Monte Carlo estimates (which is how we compute them in practice) so that $\vbhat_t \approx \sum_{\text{observed bids }x'} D_b(x',t)$
and 
$\vahat_t \approx \sum_{\text{observed asks }x'} D_a(x', t)$,
where $D_b(x,t)$, $D_a(x, t)$, and $X_t$ are given by Eq.\ 
\ref{eq:gen-mech}.
As a visual reference point, in Panel B of Fig.\ \ref{fig:orderbook} we display an example realization of Monte Carlo-approximated $\dvbhat$ and $\dvahat$, while in panel C of this figure we display the corresponding $\dx$.  

\subsection{Evolutionary dynamics}\label{sec:evo-dyn}
We provide a summary of the evolutionary dynamics in Fig.\ \ref{fig:mech-graphic}.
We first describe the selection and mutation mechanism $\mechanism$, since this mechanism changes on the objective of the simulation, and then describe the simulation in general.
We set the evolutionary mechanism $\mechanism$ to be either tournament selection-based or the identity (no evolution):
we use the tournament selection mechanism when we are attempting to evolve agents of high fitness, while we use the identity mechanism when we are generating empirical market statistics for use with real data, as described in Sec.\ \ref{sec:trading-algo-description}.

The tournament selection mechanism is standard 
\cite{blickle1996comparison}, designed as follows: given a population of evolvable agents, at the end of each generation a tournament of $\tau$ agents is selected from the population at random. We set $\tau = 17$.
These agents are sorted by fitness---their total profit $\pi$ in the market simulation of the past generation---so that
$\pi_{(1)} \geq \pi_{(2)}\geq \cdots \geq \pi_{(\tau)}$.
Agent $(i)$ is selected to remain with probability 
$p(1 - p)^{i-1}$, where we set $p = \frac{1}{2}$; 
the remaining agents are discarded. 
A total of $\tau - 1$ new agents are initialized with the parameters from the selected agent $(i)$, denoted by 
$\theta_{(i)} = (\theta_{(i),1},...,\theta_{(i),L})$ where the agent has a total of $L$ parameters. 
These new parameters are then subjected to centered Gaussian mutation; the parameters of new agent $i'$ are given by 
$\theta_{i'} = \theta_{(i)} + z_{i'}$, where 
$z_{i'} \sim \mathcal{N}(0, \gamma^2\Sigma)$ and we set $\gamma = 0.1$.
The covariance matrix $\Sigma$ of the Gaussian is diagonal, with
$\Sigma_{\ell\ell} = \text{Var}(\theta_{(i), \ell})$.
The $\tau - 1$ agents are added back into the entire population of evolvable agents for use in further generations, described in the next paragraph.

At each of $g = 1,...,G$ generations, we initialize $k = 1,...,K$ independent markets, which we set to $K = 24$.
We set $G = 100$.
In each market, we initialize $N_A$ agents with agent parameter vector distributed as $(\alpha_{a,k})_{a \in A} = \alpha_k \sim p(\alpha)$,
where $A$ is the set of agent types outlined in Sec.\ \ref{sec:agent-description}.
Given a drawn $\alpha_{k}$, there are $\alpha_{\zi,k}$ zero intelligence agents, $\alpha_{\mo,k}$ momentum agents, and so on. 
The probability distribution $p(\alpha)$ is a joint distribution over agent types that factors as a uniform distribution over the number of neural network agents and a multinomial distribution over the number of other agents that is dependent on the number of drawn neural network agents.
Given a number of neural network agents $\alpha_{\nn,k}$ drawn uniformly at random between $\nn_{\min}$ and $\nn_{\max}$, 
the remaining $N_A - \alpha_{\nn,k}$ are drawn from a multinomial distribution with probabilities $\rho_a = \frac{1}{6}$.
We set $\nn_{\min} = 2$ and $\nn_{\max} = 10$.
Within each market, agents interact \textit{vis-\`{a}-vis} the matching engine described in Sec.\ \ref{sec:auction}, trading for a total of $T$ timesteps within each generation. 
We set $T = 500$ and set the number of trading timesteps per day equal to 100, as outlined in the previous section. After $T$ timesteps, the population of neural networks is pooled---removed from each simulation and collated into one set---and the evolutionary mechanism $\mechanism$ is applied to all $\sum_{k=1}^K \alpha_{\nn, k}$ neural networks, generating a (partially) new population, as outlined in the previous paragraphs of this section. 
The new population of neural networks is then shuffled and divided into $K$ new independent markets according to the $\alpha_{\nn, k}$, along with new static agents again drawn from the multinomial distribution with equal probabilities, and the process begins again in generation $g + 1$.

\subsection{From evolved agent to trading algorithm}
\label{sec:trading-algo-description}
We convert evolved neural networks into executable trading strategies that we subsequently backtest on real financial data.
The major impediment to simply using the evolved networks as trading strategies is the lack of readily-available orderbook information for real financial markets: though such information is available for sale, it is prohibitively expensive to purchase 
\cite{tivnan2019fragmentation}.
Instead, we use orderbook data generated by 
$\generative(\alpha, \mechanism)$ as a surrogate for real orderbook data, which we believe to be an important input into the algorithms as orderbook data has been shown to carry non-zero information about future prices and be useful in making profitable short-term trading decisions
\cite{harris2005information,cao2009information,silaghi2005agent}.
We reason that, if $\generative(\alpha, \mechanism)$ is a good simulacrum of the true orderbook-generating process active in a real financial market, then the values of $\dvbhat$ and $\dvahat$ generated by the agent-based model, given an observed value of 
$\dx$ from a real market, should be similar to the changes in resting bid and ask volume that existed in the real market, even though we do not have access to that data.

Because of our lack of orderbook data, we must simulate 
$\dvbhat$ and $\dvahat$ that correspond with the observed change in price $\dx$.
To do this, we first simply draw many values from $\generative(\alpha, \text{id})$ (the generative model with no evolutionary mechanism).
Then,
given $\dx$ from the asset market, we draw multiple ($\dvbhat$, $\dvahat$) pairs
from their empirical joint pdf conditioned on this observation,
which is generated by the draws from $\generative(\alpha, \text{id})$:
\begin{equation}\label{eq:empirical-joint-conditional}
	(\dvbhat, \dvahat) \sim 
	\phat_{\generative(\alpha, \text{id})}
	(\dvb, \dva|\dx),
\end{equation}
and evaluate $f$ using the conditional expectation of these values;
this new ``marginalized'' algorithm is given by 
\begin{equation}
	\begin{aligned}
		&\fmarg( \dx_t, \dcash_t, \dshares_t, \dprofit_t)\\
		&\quad=
	 f(\dx, E[\dvbhat|\dx], E[\dvahat|\dx], \dcash, \dshares, \dprofit),
	\end{aligned}
\end{equation}
where the expectations are taken under the pdf displayed in 
Eq.\ \ref{eq:empirical-joint-conditional}.
As with the non-marginalized algorithm $f$, $\fmarg$ returns a side, change in shares, and change in price:\\
$
(\side_t, \ordershares_t, \dprice_t) = 
		f_{\text{marg}}(\dx_t, \dcash_t, \dshares_t, \dprofit_t)$.
Since we are backtesting the trading algorithm and hence it is
impossible to perform price discovery, we do not use
$\dprice_t$.
We simply simulate the execution of a market buy ($s_t = 1$) or sell ($s_t = -1$) order
for $\ordershares_t$ spot contracts of the asset and then update the algorithm's internal state.
As in the evolutionary process within the agent-based model, we allow the algorithm to sell short so that 
$\shares_t$ may be negative. 

We implement three basic risk management routines that supervise the execution of the algorithm.
These routines act as ``circuit breakers'' to halt the algorithm's
operation if certain risk limits are reached and consist of 
two versions of the traders' adage ``cut your losses but let your winners run,'' ensuring that maximum 
loss is capped at some user-set limit, 
and one leverage limit routine that halts execution if the algorithm is long or short a certain large number of 
contracts (we set this number equal to 150) \cite{aspara2015cut}.
(The interested reader is referred to the supplementary information for more detail.)
Though these routines pale in comparison with real risk-management software used in algorithmic trading 
\cite{clark2010controlling,sornette2011crashes,donefer2010algos}, 
we believe that they are sufficient for the purposes of this work.

\section{Results}\label{sec:results}

\subsection{Evolutionary dynamics}

We ran 100 independent simulations of the entire process outlined in Sec.\ \ref{sec:evo-dyn}.
This resulted in a total of 240,000 ($=$ 24 independent markets per generation $\times$ 100 generations $\times$ 100 independent simulations) conditionally independent market simulations from which to sample evolved neural networks for validation and testing on real financial asset data.
At each generation of each simulation, we saved the best individual for possible further use.

\begin{figure}
    \centering
    \includegraphics[width=\columnwidth]{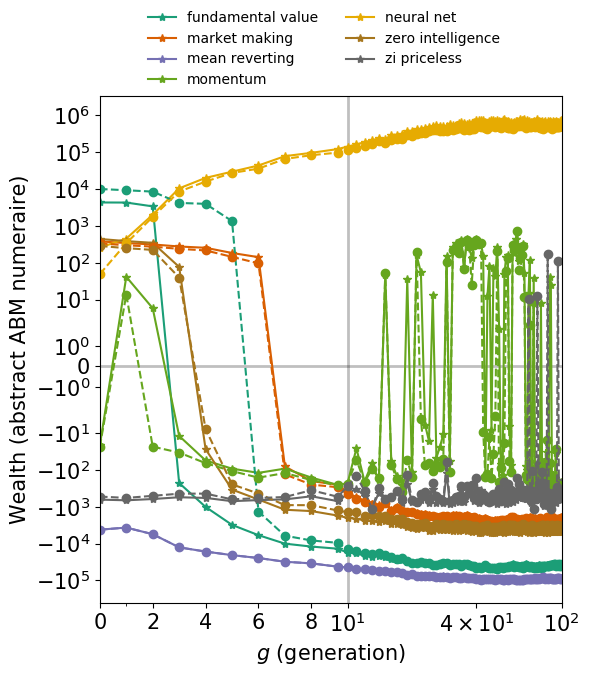}
    \caption{
    Evolving neural network agents quickly dominate all static agents; average wealth of neural network agents increases until about $g = 40$, where it plateaus while the average wealth of other static trading agents remains largely flat or decreases over time.
    In particular, mean-reverting and fundamental-value traders suffer large average wealth losses, even though fundamental-value traders start as the most profitable agent type.
    We also find that a static momentum trading strategy is, on average, the strategy that is least dominated by evolving neural networks and can actually be sustainably profitable for multiple generations; this result corresponds with the finding that a momentum-based strategy can be profitable over nontrivial timescales in real financial markets \cite{jegadeesh2001profitability}.
    }
    \label{fig:wealth-montage}
\end{figure}

Evolved neural networks ($\nn$) quickly became the dominant species of trading agent, though they were not profitable at $g = 0$ (at which point they were simply random neural networks with normally-distributed weights and biases).
Fig.\ \ref{fig:wealth-montage} displays average (solid curves) and median (dashed curves) wealth trajectories for each agent type; these statistics are calculated over all simulations at that generation.
The average and median wealth of $\nn$ agents increases quickly until about $g = 10$. It then increases more slowly until about $g = 40$, when it plateaus.
Concomitant with the rise in average and median $\nn$ wealth is a decline in the wealth of nearly every other agent type. 
In particular, though fundamental value ($\fv$) agents started out as the most profitable agent type (due to their strong beliefs about ``true'' asset values and market power; in early generations they had the ability to collectively set market price), they became the second-worst performing agent type as $\nn$ agents became more profitable.
It is notable that momentum trading agents are the only static agent type that was still able to make positive profits during multiple sequential later generations (in particular $g > 30$), long after all other static agents had become, on average, very unprofitable.
This is consistent with the finding that real asset prices may exhibit momentum effects and that trading strategies based on exploiting this momentum may result in positive expected profit
\cite{jegadeesh1993returns}.

\begin{figure}
    \centering
    \includegraphics[width=\columnwidth]{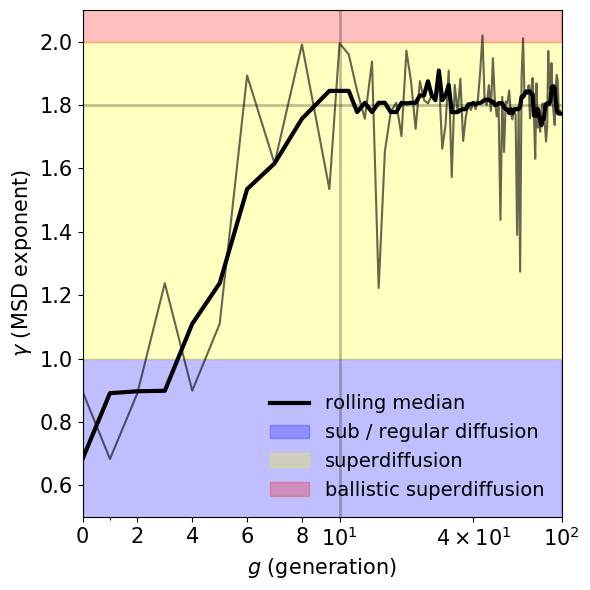}
    \caption{
    Asset price superdiffusion emerges as a byproduct of evolutionary pressure.
    Superdiffusion is defined by a superlinear relationship between mean squared deviation of a time series and time itself.
    At each generation $g$, we fit a model of the form $E[(X_t^{(g)} - \mu_t)^2]
    \propto t^{\gamma_g}$ and plot the resulting $\gamma_g$ as a function of 
    generation $g$.
    This exponent of dispersion stabilizes at roughly $\gamma_g \simeq 1.8$ after
    approximately 10 generations of evolution (indicated by the vertical black line at $g = 10$), which influences our selection of $g = 10$ for validation and testing of evolved strategies on real data.
    }
    \label{fig:price-msd}
\end{figure}

As evolution progressed through generations, statistical properties of asset price time series $X_t$ changed significantly.
The mean square deviation (MSD) of $X_t$ in generation $g$, defined by the exponent $\gamma_g$ in the relationship
$E_g[(X_t^{(g)} - \mu_t)^2] \propto t^{\gamma_g}$ where $\mu_t$ is the intertemporal mean of $X_t^{(g)}$, begins in the sub- or normally-diffusive region ($\gamma_g \leq 1$) but quickly rises to 
$\gamma_g \approx 1.8$ near $g = 10$ and remains there for the remainder of evolutionary time.
We display the MSD of asset prices by generation in Fig.\ \ref{fig:price-msd}.
MSD that grows superlinearly with time is termed anomalous superdiffusion \cite{ponzi2000evolutionary,bartolozzi2004stochastic}
and is commonly observed in real asset markets \cite{michael2003financial,cortines2007non,devi2017financial}.
This may provide evidence that observed asset price superdiffusion in real asset markets is partially driven by purely endogenous evolutionary dynamics.

\subsection{Validation and testing of evolved strategies}

We chose a subsample of the neural networks that we extracted from the market simulations for consideration as algorithmic trading strategies to be used on real data. 
Though all evolved neural networks that we saved had high fitness in the context of the agent-based model, we hypothesized that it would not be the case that all of them would have high fitness when backtested on observed asset price data.
We selected the high-fitness neural networks saved at generation $g = 10$, the set of which we will denote by $\agent_{10}$, for two reasons. 
First, this was the approximate ``elbow'' of $\log_{10} \pi$, as displayed in Fig.\ \ref{fig:wealth-montage}; at generations later than approximately $g = 10$, $\nn$ agents exhibited decreasing marginal $\log_{10} \pi$.
Second, this generation was the point at which the exponent of the MSD of asset prices appeared to stabilize at $\gamma_g \approx 1.8$.

The asset prices produced by the interaction of agents in the ABM do not appear to exhibit geometric (multiplicative noise) dynamics, but rather arithmetic (additive noise) dynamics.
Though many real financial assets do exhibit geometric- or geometric-like dynamics \cite{black1973pricing}, other assets, such as foreign exchange (FX) spot contracts, typically display quasi-arithmetic dynamics instead
\cite{hilliard1991currency}.
We thus test our evolved neural networks on FX spot rate data, eight and a half (January 2015 through July 2019) years of millisecond-sampled EUR/USD and GBP/USD spot exchange rate data sourced from an over-the-counter trading venue \footnote{The data is available from 
\href{https://www.truefx.com/}{https://www.truefx.com/}, which sources it from the Integral OCX ECN.}.
We do not implement transaction costs in our backtesting as, if these are constant, their marginal incidence on profit decreases as the amount of leverage (net number of contracts traded) increases.
We separate this dataset into a validation (2010 - 2015) and testing (2015 - 2019) split. Although this split is unnecessary in the context of overfitting to data (because the neural networks are just static at this point; their evolution occurred in the ABM and they do not update their weights based on the observed FX data), it is part of a method by which we lower the probability of false positive discovery of high-performance trading algorithms. 
The top $k$ algorithms in $\agent_{10}$, ranked by total profit accumulated by trading on validation data, were then used to trade on the test dataset; if these algorithms accumulated high profit on the validation dataset by chance, it is likely that they would not perform well on the test dataset.
We set $k = 5$ and will denote these ``elite'' trading strategies by $\agent_{\text{elite}}$.

\begin{figure}
    \centering
    \includegraphics[width=\columnwidth]{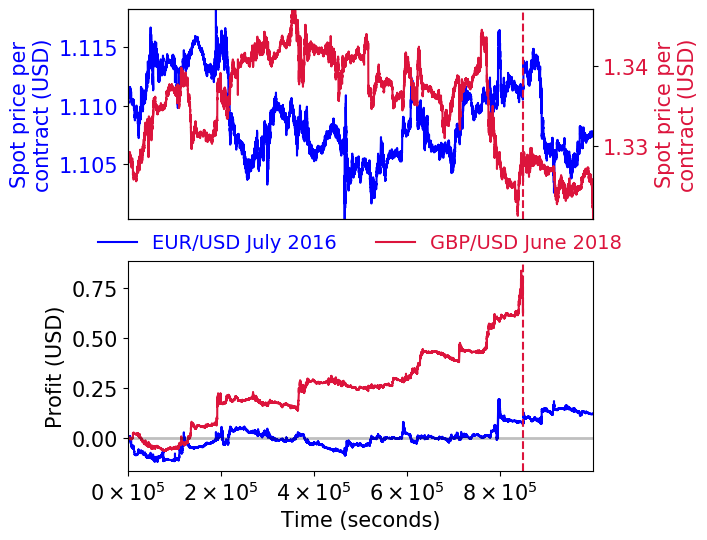}
    \caption{
Elite evolved trading algorithms are able to obtain positive profit under a wide variety of backtested trading conditions.
These price time series display both large increases and decreases during this time period, as well as regions of relatively low and high volatility.
Despite these varied conditions, an elite evolved algorithm was able to capture positive profit (shown in the blue curve) over this time period, showing large gains in profit during both price drawdowns and ramp-ups.
The vertical dashed line in each subplot indicates that the risk management system halted the execution of the trading algorithm.
In this case, this occurred because the algorithm attempted to exceed the user-set leverage limit.
    }
    \label{fig:backtest-ts}
\end{figure}

To create trading algorithms from the evolved neural networks, we followed the procedure described in Sec.\ \ref{sec:trading-algo-description}.
We resampled the spot FX rate time series at the 10s resolution, setting as $X_t$ the mean price during that 10$s$ interval multiplicatively rescaled by a constant ($100 / X_0$) so that the price on the first second of each month was equal to 100. Agents traded during the first $10^6$ seconds (approximately 16.2 days\footnote{
16.2 days $\approx \frac{10^6 \text{s}}{60\text{s/m}\times 60\text{m/h}
\times {24 \text{h/trading day}} \times \frac{5 \text{trading days}}{7\text{days}}}$
}) of each month, a number of timesteps that we chose arbitrarily to avoid possible edge effects occurring at the end of the month.
We will refer to one $10^6$s time interval of trading on a single spot rate (EUR/USD or GBP/USD) as a single trading episode.
In Fig.\ \ref{fig:backtest-ts}, we display an example time series of spot FX rate (EUR/USD during July of 2016) and corresponding profit made by an evolved neural network in this trading episode.
In this example, though the spot rate fluctuates considerably and has $|X_T - X_0| < 0.01\  \text{USD/contract}$, the profit time series increases fairly steadily throughout the entire time period, netting a total of $\pi_T \approx 0.125\  \text{USD}$. 

\begin{figure*}
    \centering
    \includegraphics[width=\textwidth]{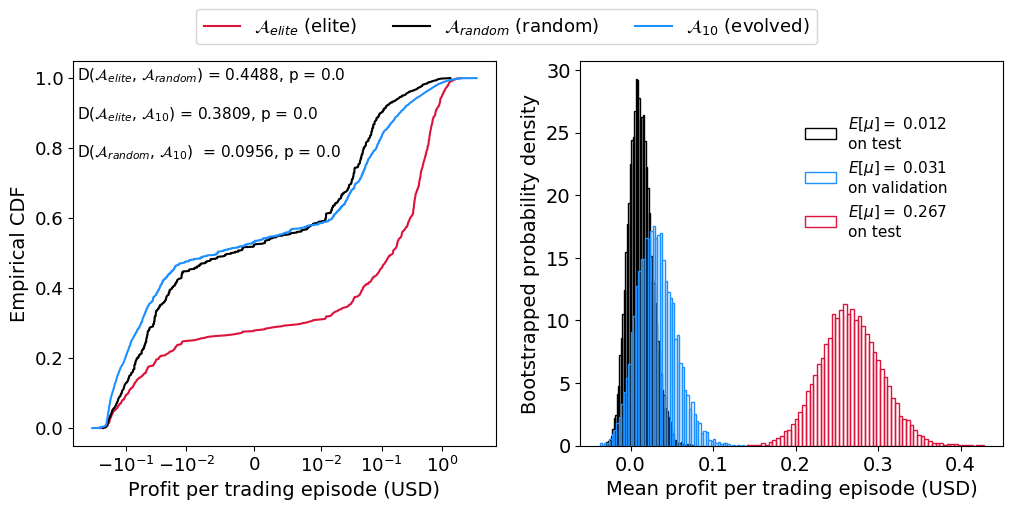}
    \caption{
The profit distributions of all evolved neural networks
($\agent_{10}$), 
random neural networks ($\agent_{\text{random}}$), 
and elite individuals ($\agent_{\text{elite}}$) 
when evaluated on real FX spot rate data differ significantly, as we demonstrate in panel A.
In panel B, we demonstrate that elite evolved neural network trading strategies have significantly higher mean profits on test data than do random neural network strategies or the set of all evolved strategies evaluated on validation data.
(The separation between validation and test data is irrelevant for the set of all evolved neural networks, as these networks evolved in the agent-based model, not through evaluation on real data.)
The data do not appear to have a diverging second moment;
we do not concern ourselves with issues that arise with bootstrapping in distributions with tail exponent $\alpha < 2$ \cite{athreya1987bootstrap}.
    }
    \label{fig:backtest-montage}
\end{figure*}
Evolved strategies differed significantly from random neural network strategies.
We compared the distribution of profit on the validation dataset by $\agent_{10}$, 
profit on the test dataset by $\agent_{\text{elite}}$, and profit on the test dataset by 20 random neural network agents ($\agent_{\text{random}}$) and display empirical cdfs of these distributions, along with bootstrapped pdfs of the means of these distributions, in Fig.\ \ref{fig:backtest-montage}.
Elite individuals (the top $k=5$ performers on the validation dataset) have the greatest maximum absolute difference (Kolmogorov-Smirnov statistic) between the empirical cdf of their profit and the other cdfs ($D(\agent_{\text{elite}}, \agent_{\text{random}})$ = 0.4488,
$D(\agent_{\text{elite}}, \agent_{10})$ = 0.3809) while the maximum distance between the empirical cdf of random neural network profit and all evolved neural network profit was smaller, but still significantly greater than zero 
($D( \agent_{10}, \agent_{\text{random}} )$ = 0.0956), as demonstrated in panel A.
Though the $\agent_{10}$ and $\agent_{\text{random}}$ profit distributions are far more similar to each other than they are to that of $\agent_{\text{elite}}$, they differ significantly in their tails: the $\agent_{10}$ distribution has a higher likelihood of observing larger losses (though the magnitude of these losses are still severely damped by the risk management routines detailed in Sec.\ \ref{sec:trading-algo-description}) and larger gains than does the $\agent_{\text{random}}$ distribution. 
It is possible this occurred because many agents in $\agent_{10}$ had high fitness in their particular realization of
the ABM with agent concentration vector $\alpha$,
but $\alpha$ did not resemble the (effectively unobservable) makeup of agents that generated the real spot price time series.
The mean profit of $\agent_{10}$ on validation data was higher than that of $\agent_{\text{random}}$ on test data, though estimates of this mean 
(displayed in panel B of Fig.\ \ref{fig:backtest-montage}) have greater dispersion that estimates of the mean of $\agent_{10}$ profit.
It is likely (probability $\approx 0.7617$) that the true mean of the 
$\agent_{10}$ profit distribution is greater than the mean of the 
$\agent_{\text{random}}$ profit distribution.
However, distributions of estimated mean for both sets of agents do contain zero (though the estimated probabilities that the mean is greater than zero are 0.7863 for $\agent_{\text{random}}$
and 0.9241 for $\agent_{10}$).

Crucially, mean $\agent_{\text{elite}}$ profit on test data is very far from zero (approximately 0.267 USD per trading episode) and the estimated distribution of mean profit is bounded away from zero.
We discuss these results further in the supplementary information.

\section{Discussion and conclusion}\label{sec:discussion}
We construct a method to develop \textit{ab initio} trading algorithms using an agent-based model of a financial market.
We subject expressive agents to evolutionary pressure, using profit generated in the agent-based model as the fitness function. 
We then backtest high-performing agents on real financial asset price data (spot foreign exchange rates). We further tested ``elite'' evolved agents---agents that performed well on validation data (EUR/USD and GBP/USD from 2010 to 2015)---on test data, the same currency pairs but during the time interval 2016 through 2019.
We find that it is possible for evolved trading algorithms to make significantly positive backtesting profit for extended periods of time during varying market conditions, even though these evolved algorithms had never experienced real financial data during the process of evolution.
This result provides evidence that a paradigm shift in the design of automated trading algorithms---
from prediction of future market states
to, instead, closely modeling the underlying mechanisms and agents of which the market is composed---may be both feasible and profitable.

Though this result is promising, it is important to note some shortcomings of our research and avenues for further exploration.
First, and most importantly, we have not actually tested the performance of the ``elite'' evolved agents in a real market, but only backtested them on real market data. The difference between these actions is profound; the ultimate expression of confidence in a trading strategy is to use it with one's own capital and we have not yet done this \cite{sandis2014skin}. Though the elite agents are consistently profitable when backtested, this does not guarantee that they will be profitable when used to trade ``live'' in a real financial market.
If elite agents generated according to the methodology laid out in Secs.\ \ref{sec:evo-dyn} and \ref{sec:trading-algo-description} are profitable when used to trade spot contracts in foreign exchange markets, we will be more confident in stating that this methodology is a robust method of generating \textit{ab initio} trading algorithms.

Second, our implementations of multiple components of our methodology were intended as proofs-of-concept; four-layer feed-forward neural networks are decidedly not at the cutting-edge of neural network design \cite{baker2016designing}. Future work could focus on improving the expressiveness and realism of agents used in the agent-based model, modifying the evolvable neural networks to use a recurrent architecture, using different order types in the matching engine (and hence increasing dimensionality of the neural networks' action space), and in general attempting to more closely match the composition of the agent-based model with the structure of modern-day securities markets (in particular, spot FX rate markets).
We could also subject other types of agents to moderate evolutionary pressure in order to simulate the knowledge dissemination that occurs in communities of relatively static strategies
\cite{charband2016online}.

Finally, there is substantial room for more analysis of the free parameters in our agent-based model---for example, tuning the parameters of the tournament selection adaptively so that later generations do not evolve to simply exploit the structure of the agent-based model but rather continue to explore novel trading strategies.
More generally, we should improve the design of the mechanism by which we select high-performing individuals from the model to be backtested on real data. 
We have used only a heuristic measure---the apparent stabilization of the MSD exponent and decreasing marginal log profit---as indicators as from which generation we should select, and what follows this is essentially rejection sampling from the space of agents that are high-performers in the agent-based model through evaluation of these agents on validation data.
We believe that there are probably better ways to implement this step.

\vspace*{10px}  
\section*{Acknowledgements}

	The authors are grateful for helpful conversations with Josh Bongard, Christopher Danforth,
	Peter Dodds, Sam Kriegman, Josh Minot, and Colin Van Oort.
	The authors are also thankful for the financial and physical support of the Massachusetts Mutual Life Insurance Company.

\bibliographystyle{unsrt}
\bibliography{sample-bibliography}

\begin{thebibliography}{10}

\bibitem{mnih2015human}
Volodymyr Mnih, Koray Kavukcuoglu, David Silver, Andrei~A Rusu, Joel Veness,
  Marc~G Bellemare, Alex Graves, Martin Riedmiller, Andreas~K Fidjeland, Georg
  Ostrovski, et~al.
\newblock Human-level control through deep reinforcement learning.
\newblock {\em Nature}, 518(7540):529, 2015.

\bibitem{silver2017mastering}
David Silver, Julian Schrittwieser, Karen Simonyan, Ioannis Antonoglou, Aja
  Huang, Arthur Guez, Thomas Hubert, Lucas Baker, Matthew Lai, Adrian Bolton,
  et~al.
\newblock Mastering the game of go without human knowledge.
\newblock {\em Nature}, 550(7676):354, 2017.

\bibitem{silver2018general}
David Silver, Thomas Hubert, Julian Schrittwieser, Ioannis Antonoglou, Matthew
  Lai, Arthur Guez, Marc Lanctot, Laurent Sifre, Dharshan Kumaran, Thore
  Graepel, et~al.
\newblock A general reinforcement learning algorithm that masters chess, shogi,
  and go through self-play.
\newblock {\em Science}, 362(6419):1140--1144, 2018.

\bibitem{chen2009co}
Shu-Heng Chen, Ren-Jie Zeng, and Tina Yu.
\newblock Co-evolving trading strategies to analyze bounded rationality in
  double auction markets.
\newblock In {\em Genetic programming theory and practice VI}, pages 1--19.
  Springer, 2009.

\bibitem{garcia2015social}
David Garcia and Frank Schweitzer.
\newblock Social signals and algorithmic trading of bitcoin.
\newblock {\em Royal Society open science}, 2(9):150288, 2015.

\bibitem{golub2018alpha}
Anton Golub, James~B Glattfelder, and Richard~B Olsen.
\newblock The alpha engine: designing an automated trading algorithm.
\newblock In {\em High-Performance Computing in Finance}, pages 49--76. Chapman
  and Hall/CRC, 2018.

\bibitem{greenwald20032002}
Amy Greenwald.
\newblock The 2002 trading agent competition: An overview of agent strategies.
\newblock {\em AI Magazine}, 24(1):83--83, 2003.

\bibitem{davidson2012dark}
Clive Davidson.
\newblock A dark knight for algos.
\newblock {\em Risk}, 25(9):32, 2012.

\bibitem{kirilenko2013moore}
Andrei~A Kirilenko and Andrew~W Lo.
\newblock Moore's law versus murphy's law: Algorithmic trading and its
  discontents.
\newblock {\em Journal of Economic Perspectives}, 27(2):51--72, 2013.

\bibitem{kissell2005understanding}
Robert Kissell and Roberto Malamut.
\newblock Understanding the profit and loss distribution of trading algorithms.
\newblock {\em Trading}, 2005(1):41--49, 2005.

\bibitem{dempster2001real}
Michael~AH Dempster and Chris~M Jones.
\newblock A real-time adaptive trading system using genetic programming.
\newblock {\em Quantitative Finance}, 1(4):397--413, 2001.

\bibitem{austin2004adaptive}
Mark~P Austin, Graham Bates, Michael~AH Dempster~3, Vasco Leemans, and Stacy~N
  Williams.
\newblock Adaptive systems for foreign exchange trading.
\newblock {\em Quantitative Finance}, 4(4):37--45, 2004.

\bibitem{lebaron1999time}
Blake LeBaron, W~Brian Arthur, and Richard Palmer.
\newblock Time series properties of an artificial stock market.
\newblock {\em Journal of Economic Dynamics and control}, 23(9-10):1487--1516,
  1999.

\bibitem{palmer1999artificial}
RG~Palmer, W~Brian Arthur, John~H Holland, and Blake LeBaron.
\newblock An artificial stock market.
\newblock {\em Artificial Life and Robotics}, 3(1):27--31, 1999.

\bibitem{lebaron2002building}
Blake LeBaron.
\newblock Building the santa fe artificial stock market.
\newblock {\em Physica A}, 2002.

\bibitem{chen2001evolving}
Shu-Heng Chen and Chia-Hsuan Yeh.
\newblock Evolving traders and the business school with genetic programming: A
  new architecture of the agent-based artificial stock market.
\newblock {\em Journal of Economic Dynamics and Control}, 25(3-4):363--393,
  2001.

\bibitem{cincotti2003wins}
Silvano Cincotti, Sergio~M Focardi, Michele Marchesi, and Marco Raberto.
\newblock Who wins? study of long-run trader survival in an artificial stock
  market.
\newblock {\em Physica A: Statistical Mechanics and its Applications},
  324(1-2):227--233, 2003.

\bibitem{cont2007volatility}
Rama Cont.
\newblock Volatility clustering in financial markets: empirical facts and
  agent-based models.
\newblock In {\em Long memory in economics}, pages 289--309. Springer, 2007.

\bibitem{martinez2009heterogeneous}
Serafin Martinez-Jaramillo and Edward~PK Tsang.
\newblock An heterogeneous, endogenous and coevolutionary gp-based financial
  market.
\newblock {\em IEEE Transactions on Evolutionary Computation}, 13(1):33--55,
  2009.

\bibitem{runarsson2005coevolution}
Thomas~Philip Runarsson and Simon~M Lucas.
\newblock Coevolution versus self-play temporal difference learning for
  acquiring position evaluation in small-board go.
\newblock {\em IEEE Transactions on Evolutionary Computation}, 9(6):628--640,
  2005.

\bibitem{akchurina2009multiagent}
Natalia Akchurina.
\newblock Multiagent reinforcement learning: algorithm converging to nash
  equilibrium in general-sum discounted stochastic games.
\newblock In {\em Proceedings of The 8th International Conference on Autonomous
  Agents and Multiagent Systems-Volume 2}, pages 725--732. Citeseer, 2009.

\bibitem{heinrich2015fictitious}
Johannes Heinrich, Marc Lanctot, and David Silver.
\newblock Fictitious self-play in extensive-form games.
\newblock In {\em International Conference on Machine Learning}, pages
  805--813, 2015.

\bibitem{Note1}
All source used in this project is available from the authors upon reasonable
  request.

\bibitem{budish2014implementation}
Eric Budish, Peter Cramton, and John Shim.
\newblock Implementation details for frequent batch auctions: Slowing down
  markets to the blink of an eye.
\newblock {\em American Economic Review}, 104(5):418--24, 2014.

\bibitem{budish2015high}
Eric Budish, Peter Cramton, and John Shim.
\newblock The high-frequency trading arms race: Frequent batch auctions as a
  market design response.
\newblock {\em The Quarterly Journal of Economics}, 130(4):1547--1621, 2015.

\bibitem{milgrom1982theory}
Paul~R Milgrom and Robert~J Weber.
\newblock A theory of auctions and competitive bidding.
\newblock {\em Econometrica: Journal of the Econometric Society}, pages
  1089--1122, 1982.

\bibitem{mcafee1987auctions}
R~Preston McAfee and John McMillan.
\newblock Auctions and bidding.
\newblock {\em Journal of economic literature}, 25(2):699--738, 1987.

\bibitem{ausubel2004efficient}
Lawrence~M Ausubel.
\newblock An efficient ascending-bid auction for multiple objects.
\newblock {\em American Economic Review}, 94(5):1452--1475, 2004.

\bibitem{Note2}
The price discovery algorithm used in our matching engine implementation is a
  modified version of the logic used in the Australian Securities Exchange's
  matching engine \cite {comerton2006influence}. An open-source implementation
  of the matching engine is at \protect \href
  {https://gitlab.com/daviddewhurst/verdantcurve}{https://gitlab.com/daviddewhurst/verdantcurve}.

\bibitem{gode1993allocative}
Dhananjay~K Gode and Shyam Sunder.
\newblock Allocative efficiency of markets with zero-intelligence traders:
  Market as a partial substitute for individual rationality.
\newblock {\em Journal of political economy}, 101(1):119--137, 1993.

\bibitem{farmer2005predictive}
J~Doyne Farmer, Paolo Patelli, and Ilija~I Zovko.
\newblock The predictive power of zero intelligence in financial markets.
\newblock {\em Proceedings of the National Academy of Sciences},
  102(6):2254--2259, 2005.

\bibitem{chan1996momentum}
Louis~KC Chan, Narasimhan Jegadeesh, and Josef Lakonishok.
\newblock Momentum strategies.
\newblock {\em The Journal of Finance}, 51(5):1681--1713, 1996.

\bibitem{badrinath2002momentum}
Swaminathan~G Badrinath and Sunil Wahal.
\newblock Momentum trading by institutions.
\newblock {\em The Journal of Finance}, 57(6):2449--2478, 2002.

\bibitem{de1989anomalies}
Werner~FM De~Bondt and Richard~H Thaler.
\newblock Anomalies: A mean-reverting walk down wall street.
\newblock {\em Journal of Economic Perspectives}, 3(1):189--202, 1989.

\bibitem{zhang2008trading}
Hanqin Zhang and Qing Zhang.
\newblock Trading a mean-reverting asset: Buy low and sell high.
\newblock {\em Automatica}, 44(6):1511--1518, 2008.

\bibitem{o1986microeconomics}
Maureen O'Hara and George~S Oldfield.
\newblock The microeconomics of market making.
\newblock {\em Journal of Financial and Quantitative analysis}, 21(4):361--376,
  1986.

\bibitem{das2008effects}
Sanmay Das.
\newblock The effects of market-making on price dynamics.
\newblock In {\em Proceedings of the 7th international joint conference on
  Autonomous agents and multiagent systems-Volume 2}, pages 887--894.
  International Foundation for Autonomous Agents and Multiagent Systems, 2008.

\bibitem{kirilenko2017flash}
Andrei Kirilenko, Albert~S Kyle, Mehrdad Samadi, and Tugkan Tuzun.
\newblock The flash crash: High-frequency trading in an electronic market.
\newblock {\em The Journal of Finance}, 72(3):967--998, 2017.

\bibitem{aldrich2018experiments}
Eric~M Aldrich and K~L{\'o}pez~Vargas.
\newblock Experiments in high-frequency trading: Testing the frequent batch
  auction.
\newblock {\em Experimental Economics}, 2018.

\bibitem{Note3}
Number of parameters is equal to number of parameters in the weight matrices
  plus the number of parameters in the bias vectors $= 350 + 33$.

\bibitem{blickle1996comparison}
Tobias Blickle and Lothar Thiele.
\newblock A comparison of selection schemes used in evolutionary algorithms.
\newblock {\em Evolutionary Computation}, 4(4):361--394, 1996.

\bibitem{tivnan2019fragmentation}
Brian~F Tivnan, David~Rushing Dewhurst, Colin~M Van~Oort, IV~Ring, H~John,
  Tyler~J Gray, Brendan~F Tivnan, Matthew~TK Koehler, Matthew~T McMahon, David
  Slater, et~al.
\newblock Fragmentation and inefficiencies in us equity markets: Evidence from
  the dow 30.
\newblock {\em arXiv preprint arXiv:1902.04690}, 2019.

\bibitem{harris2005information}
Lawrence~E Harris and Venkatesh Panchapagesan.
\newblock The information content of the limit order book: evidence from nyse
  specialist trading decisions.
\newblock {\em Journal of Financial Markets}, 8(1):25--67, 2005.

\bibitem{cao2009information}
Charles Cao, Oliver Hansch, and Xiaoxin Wang.
\newblock The information content of an open limit-order book.
\newblock {\em Journal of Futures Markets: Futures, Options, and Other
  Derivative Products}, 29(1):16--41, 2009.

\bibitem{silaghi2005agent}
Gheorghe~Cosmin Silaghi and Valentin Robu.
\newblock An agent strategy for automated stock market trading combining price
  and order book information.
\newblock In {\em 2005 ICSC Congress on Computational Intelligence Methods and
  Applications}, pages 4--pp. IEEE, 2005.

\bibitem{aspara2015cut}
Jaakko Aspara and Arvid~OI Hoffmann.
\newblock Cut your losses and let your profits run: How shifting feelings of
  personal responsibility reverses the disposition effect.
\newblock {\em Journal of Behavioral and Experimental Finance}, 8:18--24, 2015.

\bibitem{clark2010controlling}
Carol Clark et~al.
\newblock Controlling risk in a lightning-speed trading environment.
\newblock {\em Chicago Fed Letter}, 272, 2010.

\bibitem{sornette2011crashes}
Didier Sornette and Susanne von~der Becke.
\newblock Crashes and high frequency trading: An evaluation of risks posed by
  high-speed algorithmic trading.
\newblock {\em The Future of Computer Trading in Financial Markets}, 2011.

\bibitem{donefer2010algos}
Bernard~S Donefer.
\newblock Algos gone wild: Risk in the world of automated trading strategies.
\newblock {\em The Journal of Trading}, 5(2):31--34, 2010.

\bibitem{jegadeesh2001profitability}
Narasimhan Jegadeesh and Sheridan Titman.
\newblock Profitability of momentum strategies: An evaluation of alternative
  explanations.
\newblock {\em The Journal of finance}, 56(2):699--720, 2001.

\bibitem{jegadeesh1993returns}
Narasimhan Jegadeesh and Sheridan Titman.
\newblock Returns to buying winners and selling losers: Implications for stock
  market efficiency.
\newblock {\em The Journal of finance}, 48(1):65--91, 1993.

\bibitem{ponzi2000evolutionary}
Adam Ponzi and Yoji Aizawa.
\newblock Evolutionary financial market models.
\newblock {\em Physica A: Statistical Mechanics and its Applications},
  287(3-4):507--523, 2000.

\bibitem{bartolozzi2004stochastic}
Marco Bartolozzi and Anthony~William Thomas.
\newblock Stochastic cellular automata model for stock market dynamics.
\newblock {\em Physical review E}, 69(4):046112, 2004.

\bibitem{michael2003financial}
Fredrick Michael and MD~Johnson.
\newblock Financial market dynamics.
\newblock {\em Physica A: Statistical Mechanics and its Applications},
  320:525--534, 2003.

\bibitem{cortines2007non}
AAG Cortines and R~Riera.
\newblock Non-extensive behavior of a stock market index at microscopic time
  scales.
\newblock {\em Physica A: Statistical Mechanics and its Applications},
  377(1):181--192, 2007.

\bibitem{devi2017financial}
Sandhya Devi.
\newblock Financial market dynamics: superdiffusive or not?
\newblock {\em Journal of Statistical Mechanics: Theory and Experiment},
  2017(8):083207, 2017.

\bibitem{black1973pricing}
Fischer Black and Myron Scholes.
\newblock The pricing of options and corporate liabilities.
\newblock {\em Journal of political economy}, 81(3):637--654, 1973.

\bibitem{hilliard1991currency}
Jimmy~E Hilliard, Jeff Madura, and Alan~L Tucker.
\newblock Currency option pricing with stochastic domestic and foreign interest
  rates.
\newblock {\em Journal of Financial and Quantitative Analysis}, 26(2):139--151,
  1991.

\bibitem{Note4}
The data is available from \protect \href
  {https://www.truefx.com/}{https://www.truefx.com/}, which sources it from the
  Integral OCX ECN.

\bibitem{Note5}
16.2 days $\approx \protect \frac {10^6 \protect \text {s}}{60\protect \text
  {s/m}\times 60\protect \text {m/h} \times {24 \protect \text {h/trading day}}
  \times \protect \frac {5 \protect \text {trading days}}{7\protect \text
  {days}}}$.

\bibitem{athreya1987bootstrap}
KB~Athreya et~al.
\newblock Bootstrap of the mean in the infinite variance case.
\newblock {\em The Annals of Statistics}, 15(2):724--731, 1987.

\bibitem{sandis2014skin}
Constantine Sandis and Nassim Taleb.
\newblock The skin in the game heuristic for protection against tail events.
\newblock {\em Review of Behavioral Economics}, 2014.

\bibitem{baker2016designing}
Bowen Baker, Otkrist Gupta, Nikhil Naik, and Ramesh Raskar.
\newblock Designing neural network architectures using reinforcement learning.
\newblock {\em arXiv preprint arXiv:1611.02167}, 2016.

\bibitem{charband2016online}
Yeganeh Charband and Nima~Jafari Navimipour.
\newblock Online knowledge sharing mechanisms: a systematic review of the state
  of the art literature and recommendations for future research.
\newblock {\em Information Systems Frontiers}, 18(6):1131--1151, 2016.

\bibitem{comerton2006influence}
Carole Comerton-Forde and James Rydge.
\newblock The influence of call auction algorithm rules on market efficiency.
\newblock {\em Journal of Financial Markets}, 9(2):199--222, 2006.

\bibitem{dewhurst2019selection}
David~Rushing Dewhurst, Michael~Vincent Arnold, and Colin~Michael Van~Oort.
\newblock Selection mechanisms affect volatility in evolving markets.
\newblock In {\em Proceedings of the Genetic and Evolutionary Computation
  Conference}, pages 90--98. ACM, 2019.

\end{thebibliography}

\appendix 

\section{Supplementary information }

\textbf{Algorithmic details of agents}
\medskip

All agents (except for neural network agents) submit a number of shares that is Poisson distributed with mean value 
$E[N] = 100$ by default. 

\begin{itemize}
\item Zero intelligence ($\zi$): submit bid with probability $s_{t} \sim \text{Bernoulli}(\pbid)$.
We set $\pbid = \frac{1}{2}$.  
Order price is distributed uniformly around the last equilibrium price from the matching engine, 
$X_t^{(o)} = X_{t-1} + \nu u_t$, where $u_t \sim \mathcal{U}(-1, 1)$ and $\nu > 0$ is the so-called ``micro-volatility'' preference 
(denoted below by \texttt{self.vol\_pref}) of the agent
\cite{dewhurst2019selection}.
We have set $\nu \sim \text{Exponential}(\beta=10)$ in this study.
This random variable is set at runtime and remains constant for the life of the agent.
\item Zero intelligence plus ($\zip$): the same as a zero-intelligence agent except they submit no order price; their orders are market orders. 
\item Momentum ($\mo$): these agents implement a very simple momentum algorithm based on the change in price between the current price and the last observed price:
	\begin{widetext}
\begin{lstlisting}
dp = self.last - p  # p is the price; self.last is the last observed price
if dp > 0:
    side = 'ask'
    price = p + self.dx  # self.dx is a small price increment
elif dp < 0:
    side = 'bid'
    price = p - self.dx
else:
    side = np.random.choice(['ask', 'bid'])
    price = p
\end{lstlisting}
	\end{widetext}
We set the price increment $dx = 0.05$, but this is obviously arbitrary. In particular, one could set this to be a random variable that takes other information about market state as parameters.
\item Fundamental value: these agents have a ``true valuation'' estimate of what they think the traded asset is actually worth. If the price of the asset is below this value, they submit a bid, while if it is above this value, they submit an ask.
	\begin{widetext}
\begin{lstlisting}[mathescape=true]
if p == orders.NaP:  # how we denote no price information returned by matching engine
    return []
# mean_price is the agent's true-valuation estimate
# price_tolerance is the agent's estimate of error around the true mean price
# that they are willing to accept
elif p > self.mean_price + self.price_tolerance:
    side = 'ask'
    price = p + self.dx + self.vol_pref * np.random.random() - self.vol_pref / 2  # vol_pref is nu from above
elif p < self.mean_price - self.price_tolerance:
    side = 'bid'
    price = p - self.dx + self.vol_pref * np.random.random() - self.vol_pref / 2
else:
    return []
\end{lstlisting}
	\end{widetext}
As with any other agent with a \texttt{vol\_pref} attribute,
this algorithm can be made deterministic by setting \texttt{self.vol\_pref} $= 0$.

\item Mean reversion: this agent anticipates a return to some rolling mean. If the price is higher than the rolling mean, the agent submits an ask order. If the price is lower than the rolling mean, it submits a bid order.
	\begin{widetext}
\begin{lstlisting}
self.prices[self.current_ind] = p  # a list of recent prices
# self.window is number of prices that are stored
self.current_ind = (self.current_ind + 1) % self.window  

self.mean_price = self.prices[self.prices > 0].mean()
if p == orders.NaP:
    return []
elif p > self.mean_price + self.price_tolerance:
    side = 'ask'
    price = p - self.vol_pref * np.random.random()
elif p < self.mean_price - self.price_tolerance:
    side = 'bid'
    price = p + self.vol_pref * np.random.random()
else:
    return []
\end{lstlisting}
	\end{widetext}
We set \texttt{self.window} $=50$ by default.

\item Market making: this agent provides liquidity on both sides of a limit order book in an attempt to collect small profits that result from other market participants crossing the spread.
Like the mean reverting agent, this agent keeps a rolling list of recent prices to which it refers when calculating likelihood of price movements.
\begin{widetext}
\begin{lstlisting}
self.prices[self.current_ind] = self.engines[eid].eq  # interfacing with one of possibly many matching engines
self.current_ind = (self.current_ind + 1) % self.window
p = self.prices[self.prices > 0].mean()

# React to inventory imbalance
# target_inventory_size is generally assumed to be a small number
divergence = (self.shares_held - self.target_inventory_size)
if divergence > self.inventory_tolerance:
    self.shift = max(0.01, self.shift - 0.01)  # shifts reference price
    self.spread += 0.01  # shifts how far on either side of the equilibrium the agent will place orders
elif divergence < -self.inventory_tolerance:
    self.shift += 0.01  # shift price up
    self.spread += 0.01
else:
    self.shift = 0.
    self.spread = max(0.01, self.spread - 0.01)
\end{lstlisting}
\end{widetext}
Unlike the other agents described, the market-making agent then submits two orders instead of only one.
\begin{widetext}
\begin{lstlisting}
buy_order = orders.Order(
    self.uid,
    f'{self.uid}-{self.order_number}',
    'bid',
    max(int(self.shares - divergence), 10),  # adaptively calculate number of shares to lower inventory imbalance
    np.round(p - self.spread + self.shift, 2),
    time_in_force=0,
)
self.order_number += 1

sell_order = orders.Order(
    self.uid,
    f'{self.uid}-{self.order_number}',
    'ask', max(int(self.shares + divergence), 10),
    np.round(p + self.spread + self.shift, 2),
    time_in_force=0,
)
self.order_number += 1
\end{lstlisting}
\end{widetext}
\end{itemize}
\medskip

\noindent
\textbf{Marginalization procedure}
\medskip

We provide more details on how we attempt to marginalize over unobserved 
market variables using analogues of those variables generated by the agent-based model (ABM).
Recall from the main paper that we can view the ABM as a stochastic function
$\generative(\alpha, \mechanism)$ 
that, given an agent parameter vector $\alpha$ and evolutionary mechanism $\mechanism$ (which may be the identity mechanism, i.e.\ no selection or mutation),
generates orderbooks $D_b(x,t)|\alpha$, $D_a(x,t)|\alpha$ and price time series $X_t|\alpha$. Our first step is simply to jointly obtain many orderbooks and price time series by calling $\generative(\alpha, \mechanism)$ many times for a variety of different $\alpha$.
We then have random fields of orderbooks
$\mathcal{D}_b = \{ D_b(x,t)|\alpha \}_{\alpha}$,
$\mathcal{D}_a = \{ D_a(x,t)|\alpha \}_\alpha$ and 
price time series $\mathcal{X} = \{ X_t|\alpha \}_\alpha$ indexed by the vectors $\alpha$ that we draw from some joint distribution $p(\alpha)$.
In our implementation, we just set $p(\alpha)$ to a multinomial distribution with equal probabilities of picking each agent type. 
\medskip

\noindent
We called $\generative(\alpha, \text{id})$ once for each saved evolved neural network and added the evolved neural network to the simulation. 
Given these random fields of orderbooks and price time series, we create the price difference $\dx = X_t - X_{t-1}$ and the total resting order volume time series calculated from the sequence of orderbooks,
$\dvbhat$ and $\dvahat$, defined analogously.
The change in price does have an observable real market analogue---the actual change in price of the spot asset---while the change in resting order volume does not have an observable real market analogue, since we did not have access to real market orderbook data.
We store $\dx$ in a ball tree using the $\ell_1$ distance metric.
Given an observed $\dx^*$ from real market data, we then query the ball tree for 
the indices $i_1,...,i_k$ of the nearest $k$ neighbors of $\dx^*$, 
$\dx_{i_1},...,\dx_{i_k}$ and then extract
$\dvbhat_{i_1},...,\dvbhat_{i_k}$ and $\dvahat_{i_1},...,\dvahat_{i_k}$.
We then approximate the market $(\dvb, \dva)|\dx^*$ by the approximation to the conditional expectation given by the result of the nearest-neighbors query:
\begin{align}
	\dvb|\dx^* &\approx \frac{1}{k}\sum_{j=1}^k \dvbhat_{i_j},\\
	\dva|\dx^* &\approx \frac{1}{k}\sum_{j=1}^k \dvahat_{i_j}.
\end{align}
The validity of this procedure rests on the degree to which the mechanics and dynamics of the ABM simulate the true mechanics and dynamics of the real asset market.
\medskip

\noindent
\textbf{Risk management routines}
\medskip

We implemented some simple risk management routines to halt the losses of poorly-performing algorithms. 
These routines come in three variants: two versions of the traders' adage ``cut your losses but let your winners run,'' and one version that seeks to limit total leverage (net open position of spot contracts).
\begin{itemize}
    \item ``Cut losses'' algorithm operating on price levels: halts trading if total profit is below a certain level.
	    \begin{widetext}
    \begin{lstlisting}
    def cut_losses(
        profit_arr,
        lower_limit=-0.5,
        method='absolute',
        roll_ind=100,
        ):
    """Implements the level-based adage "cut your losses and let your winners run."

    If the level of profit is below a certain set level, halt trading. Otherwise, keep going.

    :param profit_arr: array of profits so far
    :type profit_arr: index-able
    :param lower_limit: the level of profit below which trading should halt.
    :param method: one of `absolute` or `rolling`, how to calculate the maximum profit setpoint
    :type lower_limit: `float`

    :returns : `bool`, whether or not trading should halt
    """
    if (method == 'absolute') or (len(profit_arr) <= roll_ind):
        if profit_arr[-1] <= lower_limit:
            return True
        return False

    elif method == 'rolling':
        max_profit = np.max( profit_arr[-roll_ind:] )
        if profit_arr[-1] - lower_limit < max_profit:
            return True
        return False

    else:  # they didn't specify anything, better to halt rather than propagate their errors
        return True
    \end{lstlisting}
	    \end{widetext}
    If \texttt{method == 'rolling'} and \texttt{roll\_ind} $=0$, this is equivalent to halting trading at time $t$ if $\pi_t$ is less than $\max_{t' = 1,...,t} \pi_{t'} - $ \texttt{lower\_limit}; we used these parameter settings during our experiments.
    \item ``Cut losses'' algorithm operating on changes in price level: halts trading if $\dprofit$ is less than some specified cutoff.
	    \begin{widetext}
    \begin{lstlisting}
    def cut_losses_delta(
        profit_arr,
        lower_limit=-0.5,
        ):
    """Implements the flow-based adage "cut your losses and let your winners run."

    If the change in profit is below a certain set level, halt trading. Otherwise, keep going.

    :param profit_arr: array of profits so far
    :type profit_arr: index-able
    :param lower_limit: the level of delta profit below which trading should halt.
    :type lower_limit: `float`

    :returns : `bool`, whether or not trading should halt

    """
    if (len(profit_arr) >= 2) and (profit_arr[-1] - profit_arr[-2] <= lower_limit):
        return True
    return False
    \end{lstlisting}
	    \end{widetext}
    \item Leverage limit: if the total number of spot contracts (long or short, i.e., positive or negative) is above a set threshold, halts trading.
	    \begin{widetext}
    \begin{lstlisting}
    def limit_leverage(
        shares_arr,
        max_shares=150,
        ):
    if np.abs( shares_arr[-1] ) >= max_shares:
        return True
    return False
    \end{lstlisting}
	    \end{widetext}
\end{itemize}
The existence of a risk management routine supervising an algorithm's execution will, in general, change the algorithm's profit distribution.
As an example, suppose that a trading algorithm had zero average profit
$E[\pi] =\int_{-\infty}^{\infty} \pi p(\pi)\ d\pi = 0$ when unsupervised by a risk management algorithm.
If the risk management algorithm were ``perfect'' in the sense that it could guarantee limit loss to no less than $-\pi^*$ for $\pi^* > 0$, then it is the case that the trading algorithm would have positive expected profit,
since expected profit is then given by 
\begin{align}
\int_{-\infty}^{\infty}
\max\{-\pi^*, \pi \}\ p(\pi)\ d\pi \label{eq:profit-option}
\geq \int_{-\infty}^{\infty} \pi\ p(\pi)\ d\pi
= 0.
\end{align}
\medskip

\noindent
\textbf{Profitability of evolved agents}\\
We tested the profitability of all evolved algorithms on the validation dataset, the currency pairs EUR/USD and GBP/USD from 1-1-2010 to 12-31-2015, and then tested the elite evolved algorithms---what we termed the top five most profitable algorithms on the validation dataset---on a test dataset, EUR/USD and GBP/USD from 1-1-2016 to 7-1-2019 (the date we collected this data).
\medskip

\noindent
Below, we display the distributions of profit by the elite evolved algorithms on the test dataset.
By total profit, we mean all profit the algorithm earned over all trading episodes (defined in the main paper) in the test dataset.
The 
probability of profit, expected profit, and maximum \textit{a-posteriori} 
profit are all measured on a per-trading-episode basis.
The red curve superimposed on the histogram is a maximum-likelihood estimated lognormal pdf, which fits the observed data fairly well in some cases and not well in others. 
We fit a lognormal pdf as, if profit accumulates via random positive and negative percentages changes, the lognormal distribution is the appropriate limiting distribution by the multiplicative CLT (after shifting the distribution by the appropriate location and scale parameters).
\begin{figure*}
    \centering
    \includegraphics[width=\textwidth]{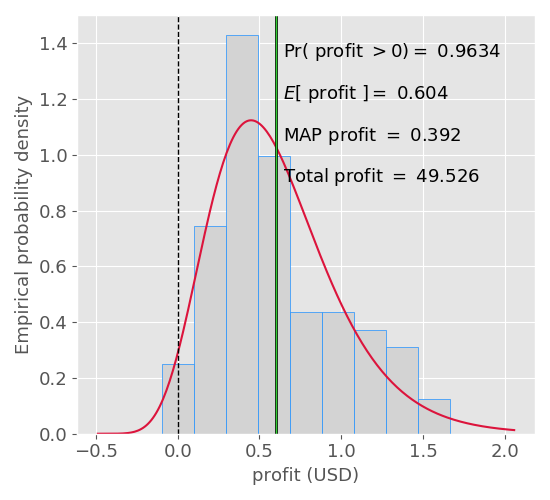}
    \caption{$g = 10$, independent simulation 21}
    \label{fig:21}
\end{figure*}
\begin{figure*}
    \centering
    \includegraphics[width=\textwidth]{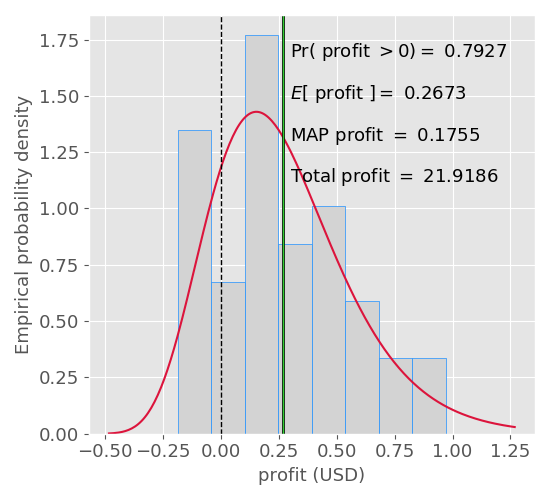}
    \caption{$g = 10$, independent simulation 1}
    \label{fig:1}
\end{figure*}
\begin{figure*}
    \centering
    \includegraphics[width=\textwidth]{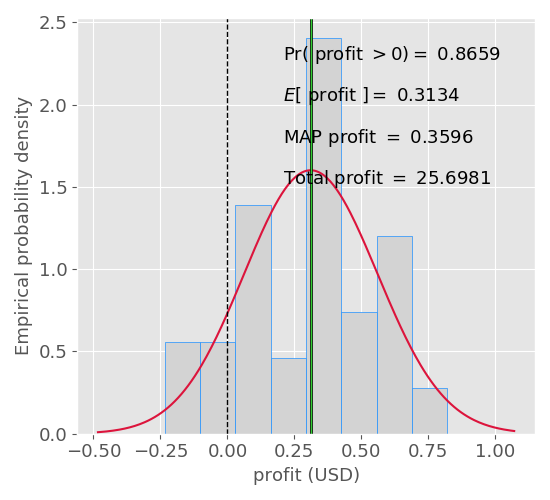}
    \caption{$g = 10$, independent simulation 93}
    \label{fig:93}
\end{figure*}
\begin{figure*}
    \centering
    \includegraphics[width=\textwidth]{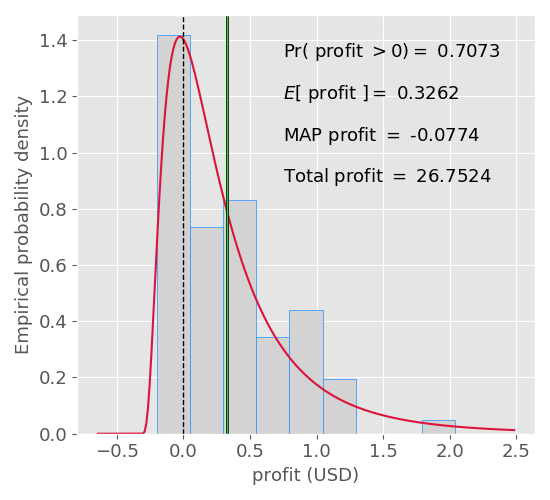}
    \caption{$g = 10$, independent simulation 79}
    \label{fig:79}
\end{figure*}
\begin{figure*}
    \centering
    \includegraphics[width=\textwidth]{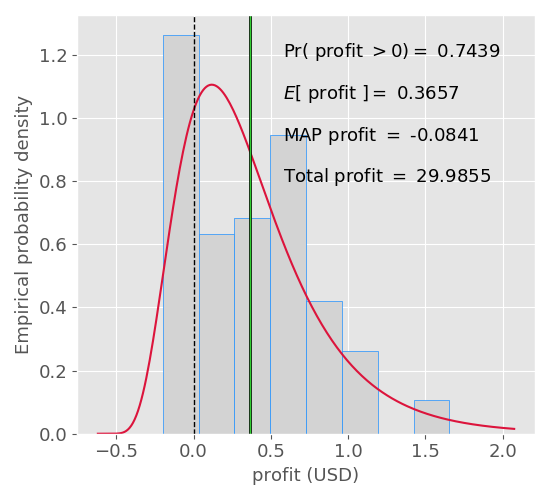}
    \caption{$g = 10$, independent simulation 38}
    \label{fig:38}
\end{figure*}

\end{document}